\documentclass[letterpaper, 10 pt, conference]{ieeeconf}  %

\IEEEoverridecommandlockouts                              %

\usepackage{times}
\usepackage{dsfont}
\usepackage{graphicx}
\usepackage{multicol}
\usepackage[bookmarks=true]{hyperref}
\hypersetup{
    colorlinks=true,
    linkcolor=blue,
    filecolor=magenta,      
    urlcolor=black, %
    pdftitle={SuReNav_ICRA_2026},
    pdfpagemode=FullScreen,
    }
\usepackage[utf8]{inputenc} %
\usepackage[T1]{fontenc}    %
\usepackage{url}            %
\usepackage{booktabs}       %
\usepackage{amsfonts}       %
\usepackage{amssymb}
\usepackage{amsmath} %
\usepackage{nicefrac}       %
\usepackage{xcolor}         %

\usepackage{caption}
\usepackage{subcaption}

\usepackage{bbm}
\usepackage{algorithm}
\usepackage{algpseudocode}
\usepackage{framed}
\usepackage{siunitx}
\usepackage{bbm}

\let\labelindent\relax
\usepackage{enumitem}

\usepackage{mathtools}
\graphicspath{ {images/} }

\usepackage{svg}
\usepackage{multirow}

\newcommand{\vA}{\mathbf{A}}
\newcommand{\vc}{\mathbf{c}}

\newcommand{\vE}{\mathbf{E}}
\newcommand{\vf}{\mathbf{f}}

\newcommand{\vr}{\mathbf{r}}

\newcommand{\vv}{\mathbf{v}}
\newcommand{\vV}{\mathbf{V}}
\newcommand{\vx}{\mathbf{x}}

\DeclareMathOperator*{\argmin}{arg\,min}

\title{\LARGE \bf
SuReNav: Superpixel Graph-based Constraint Relaxation \\for Navigation in Over-constrained Environments
}

\author{Keonyoung Koh, Moonkyeong Jung, Samuel Seungsup Lee and Daehyung Park\textsuperscript{\textdagger}%
\thanks{All authors are with the School of Computing, Korea Advanced Institute of Science and Technology, Korea ({\tt\small \{keonyoung, jmk7791, leesamuel1007, daehyung\}@kaist.ac.kr}). {\textsuperscript{\textdagger}}D. Park is the corresponding author.
This research was supported by the National Research Council  of Science \& Technology(NST) grant by the Korea government(MSIT)(No. GTL25041-000), the Institute of Information \& communications Technology Planning \& Evaluation(IITP) grant funded by the Korea government(MSIT)(No. RS-2024-00509279, RS-2024-00457882), and the Artificial intelligence industrial convergence cluster development project funded by the Ministry of Science and ICT(MSIT, Korea) \& Gwangju Metropolitan City.}
}

\begin{document}

\maketitle
\thispagestyle{empty}
\pagestyle{empty}

\begin{abstract}

We address the over-constrained planning problem in semi-static environments. The planning objective is to find a best-effort solution that avoids all hard constraint regions while minimally traversing the least risky areas. Conventional methods often rely on pre-defined area costs, limiting generalizations. Further, the spatial continuity of navigation spaces makes it difficult to identify regions that are passable without overestimation. To overcome these challenges, we propose SuReNav, a superpixel graph-based constraint relaxation and navigation method that imitates human-like safe and efficient navigation. Our framework consists of three components: 1) superpixel graph map generation with regional constraints, 2) regional-constraint relaxation using graph neural network trained on human demonstrations for safe and efficient navigation, and 3) interleaving relaxation, planning, and execution for complete navigation. We evaluate our method against state-of-the-art baselines on 2D semantic maps and 3D maps from OpenStreetMap, achieving the highest human-likeness score of complete navigation while maintaining a balanced trade-off between efficiency and safety. We finally demonstrate its scalability and generalization performance in real-world urban navigation with a quadruped robot, Spot. Code and Videos are available at https://sure-nav.github.io/.
\end{abstract}

\section{Introduction}
\begin{figure}[t]
 \centering
  \includegraphics[width=0.95\linewidth]{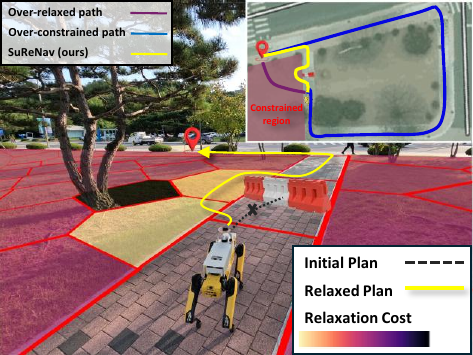}
  \caption{A capture of regional-constraint relaxation experiment in an over-constrained navigation environment. A quadruped robot pre-plans a path on a known map, and when the path is blocked, our method finds a new path by relaxing a compact area with the least risky constraints to enable safe and efficient human-like navigation.}
  \label{fig:teaser}
  \vspace{-1em}
\end{figure}

Consider the problem of over-constrained planning in semi-static environments. The planning is to find an optimal plan that satisfies all hard constraints while possibly violating less critical soft constraints. For example, a mobile robot navigating a park may encounter a blocked path on a pedestrian road (see Fig.~\ref{fig:teaser}). Although typical navigation rules dictate that the robot should not traverse the lawn, it can be more efficient to slightly deviate onto the grass rather than a long detour or backtracking. Such environmental changes frequently occur, leading to over-constrained planning in real-world settings\textemdash scenarios we define as semi-static environments.

Traditional approaches, often framed as constraint satisfaction problems, find the best solution by minimizing constraint violations. Common approaches include the minimum violation problem (MVP)~\cite{reyes2013incrementalgformvp}, minimum constraint removal~\cite{hauser2014minimum}, and violation cost estimation~\cite{ganganath2018shortest}. For instance, Kris Hauser's minimum constraint displacement method improves the current best planning solution by selecting a subset of obstacles to displace~\cite{hauser2013minimum}. Kim et al. relax a regional constraint that makes a current plan locally infeasible in temporal logic-based planning~\cite{kim2024reactive}. Despite these efforts, a major limitation is their reliance on uniform or pre-defined priorities for constraint selections, which are hard to generalize to complex environments with diverse properties of intricate navigation contexts (e.g., distinguishing between a necessary road crossing to follow a sidewalk and unsafe jaywalking). Therefore, desired approaches should relax constraints in a manner that simultaneously considers safety and efficiency. 

Another challenge is that regional constraints should accurately represent to relax necessary and not overestimated space. Typical navigation approaches distinguish passable or impassable regions only, based on visual segments~\cite{roth2024viplanner}. The representation is hard to indicate partial regional constraints, where the relaxation of entire segments, such as road or grass, potentially generates the overly relaxed shortest paths. Although pixel-based representation is another option, spatial continuity complicates the definition of \textit{unit} constraints.

In this context, we propose SuReNav, a constrained-region-aware, superpixel graph-based relaxation and navigation method. SuReNav aims to generate human-like complete navigation paths without requiring pre-defined or overestimated regional constraints. To address these, our method comprises three key components: 1) superpixel graph-map generation, 2) graph neural network (GNN)-based regional-constraint relaxation, 3) interleaving relaxation, planning, and execution. The superpixel graph models the navigation space with compact segments (i.e., superpixels) that preserve label consistency and boundaries within a graph structure. This graph representation enables the relaxation of less critical constraints by coupling a GNN-based relaxation-cost estimator with a differentiable A* planner, which propagates risks and task objectives in a manner inspired by human demonstrations. Finally, the interleaved relaxation, planning, and execution process ensures safe, efficient, and complete navigation.

We validate our method through quantitative and qualitative analyses in semi-static simulated and real-world navigation studies. For the statistical evaluation, we introduce a new navigation benchmark generating OpenStreetMap-based navigation scenarios with dynamic environmental changes reflecting semi-static conditions. Experimental results show that our proposed method outperforms four representative baselines in terms of human-likeness while achieving comparable completeness, safety, and efficiency. Finally, we deploy the system on a quadruped robot in a university campus demonstrating its scalability and generalizability under real-world operating conditions.

Our main contributions are as follows:
\begin{itemize}[leftmargin=*, noitemsep,topsep=0pt]
    \item We formalize a novel constrained-region relaxation and planning problem for reactive path planning in semi-static environments.
    \item We introduce a graph-based constrained-region relaxation cost estimator that enables safe and efficient human-like complete navigation.
    \item We propose a joint training framework that learns both the cost estimator and a neural path planner by leveraging the differential A* algorithm.
    \item We validate SuReNav through extensive evaluations in simulation and real-world navigation scenarios.    
\end{itemize}

\section{Related Work}
Traditional navigation planning methods typically require either pre-defining traversable regions~\cite{drouilly2015semanticrep} or estimating regional traversability~\cite{gupta2020cogmap} before or during navigation. While these approaches explicitly model spatial and semantic constraints, their reliance on human intervention limits generalization to novel environments, often reducing completeness or optimality~\cite{zhao2019semantic}.
Alternatively, recent end-to-end methods infer actions directly from observations without hand-crafted regional constraints by leveraging large-scale datasets~\cite{mousavian2019visual,sridhar2024nomad,chaplot2020object}. However, the data-driven mappings lack interpretable mechanisms for preserving explicit constraints during planning~\cite{mirowski2018learning}.

On the other hand, a number of navigation planners address the MVP, even in over-constrained environments, by modeling semantically recognized navigation constraints as traversability costs~\cite{reyes2013incrementalgformvp,ganganath2018shortest}.
These cost-based planners assign violation costs to individual constraints and optimize navigation by minimizing the total cost trade-offs across constraints. However, rule-based or cost metrics often fail to capture complete context awareness~\cite{kreutzmann2013safenav} and struggle with precise spatial constraint modeling~\cite{koenig2002d}.
Thus, developing robust constraint-to-cost mappings remains an open challenge.

The advent of search-guided learnable planners offers a novel direction for addressing over-constrained planning problems. Recent approaches learn costs and heuristics with neural networks~\cite{yonetani2021path,khan2021badgr}, or perform efficient searches over traversal-cost graphs generated by GNNs~\cite{zang2023graphmp}. 
By learning cost functions or heuristics from data, these methods encode context-specific risks while preserving the interpretability and completeness of search-based planning. Building on this paradigm, our model leverages data-driven learning to estimate violation costs for regional constraints defined on a superpixel-based region graph, thereby enabling constraint-aware relaxation and safe, efficient navigation in over-constrained environments. %

\section{Preliminaries}\label{sec_preliminary}
We review the \textit{differentiable graph-based A* planner} proposed by GraphMP~\cite{zang2023graphmp}, to enable end-to-end training of a relaxation cost estimator using the search result from the region--adjacency graph $G=(\mathcal{V}, \mathcal{E})$ where $\mathcal{V}$ and $\mathcal{E}$ denote the node and edge sets, respectively.

GraphMP extends Neural A* \cite{yonetani2021path}'s grid-based differentiable A* search module to general graphs. The method maintains three containers: the open-set mask $\mathbf{o}\in\mathbb{B}^{|\mathcal{V}|}$, the closed-set mask $\mathbf{c}\in\mathbb{B}^{|\mathcal{V}|}$, and the accumulated path costs $\mathbf{g}\in\mathbb{R}^{|\mathcal{V}|}$, where the entries with $1$ denote the included nodes.

At each expansion step with the open-set mask $\mathbf{o}$, the search process first forms a temperature-controlled soft priority over the open set with heuristic values $\mathbf{h}\in\mathbb{R}^{|\mathcal{V}|}$ and then returns a one-hot selection:
\begin{align}
\mathbf{v}_{\text{sel}} = I_{\text{max}}\left(\frac{\exp(-(\mathbf{g}+\mathbf{h})/\lambda)\odot\mathbf{o}}{\exp(-(\mathbf{g}+\mathbf{h})/\lambda)\mathbf{o}}\right),
\label{eq_v_sel}
\end{align}
where $\lambda>0$ is a temperature, $\odot$ denotes the Hadamard product, and $I_\text{max}$ produces a minimal-distance node mask $\mathbf{v}_\text{sel}\in\mathbb{B}^{|\mathcal{V}|}$. 

After selecting $\mathbf{v}_{\text{sel}}$, differentiable graph-based A* updates containers and identifies undiscovered neighbors $\mathbf{v}_{\text{nbr}}\in\mathbb{B}^{|\mathcal{V}|}$ using the adjacency matrix $\mathbf{A}\in\mathbb{B}^{|\mathcal{V}|\times |\mathcal{V}|}$: 
\begin{align}
\mathbf{o}\leftarrow\mathbf{o}-\mathbf{v}_{\text{sel}}, ~\mathbf{c}\leftarrow\mathbf{c}+\mathbf{v}_{\text{sel}},
\mathbf{v}_{\text{nbr}}=\mathbf{A}\mathbf{v}_{\text{sel}}\odot(\mathbf{1} -\mathbf{c}).
\end{align}
where $\mathbf{1}$ is for all-one vector. Upon the identification, the search module then computes the accumulated costs $\mathbf{g}'$ associated with the path containing the selected node $\vv_{\text{sel}}$,
\begin{align}
\mathbf{g}'&=\mathbf{g}\odot\mathbf{v}_\text{sel}+\mathbf{W}\mathbf{v}_\text{sel},
\label{eq_g_prime}
\end{align}
where a weighted adjacency matrix $\mathbf{W}\in\mathbb{R}^{|\mathcal{V}|\times |\mathcal{V}|}$. Finally, the module updates the accumulated costs $\mathbf{g}$ comparing the node-wise cost minimums and expands the open-set mask $\mathbf{o}$ including all eligible neighbors following Eq.~(7) and (8) in \cite{zang2023graphmp} (see \cite{zang2023graphmp} for the full formulation).

GraphMP jointly trains the neural heuristic estimator and the search procedure end-to-end. We exploit its graph-based gradient flow to learn the relaxation-cost estimator comparing the output plan and human demonstration paths.

\begin{figure*}[t]
    \centering
    \includegraphics[width=\textwidth]{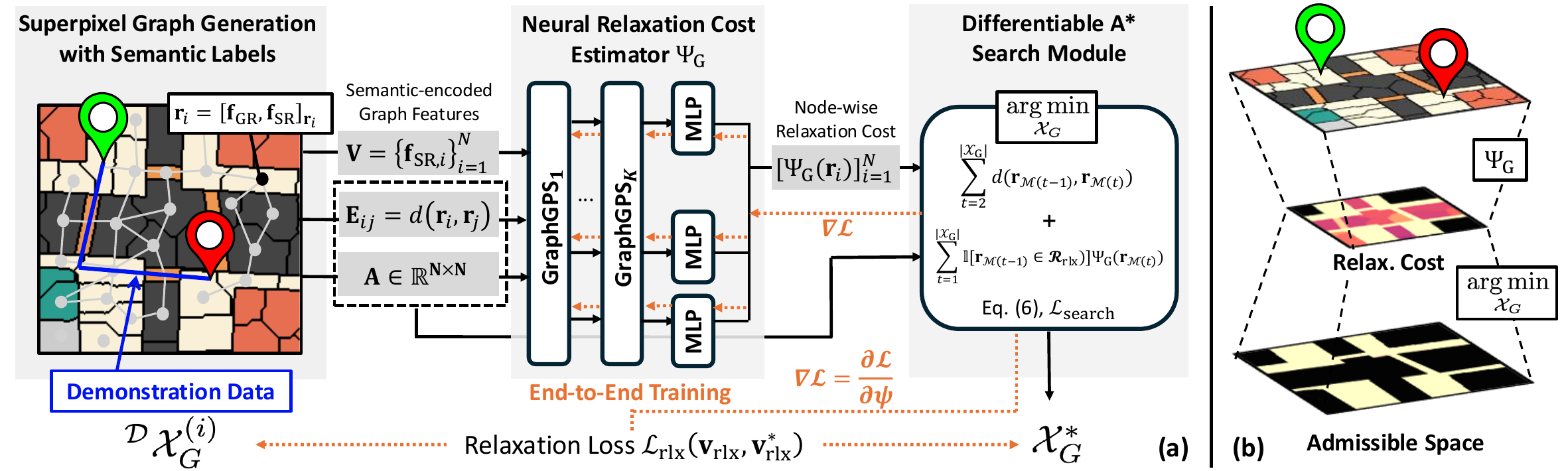}
    \caption{
    Overview architecture of SuReNav that automatically relaxes regional constraints while planning a graph path $\mathcal{X}_G$ in semi-static navigation environment. (a) In the training phase, SuReNav generates a superpixel graph from a 2-dimensional map with safety-relevant semantic features. A neural relaxation cost estimator $\Psi_G$ then computes a node-wise relaxation cost, which is incorporated as an optimization term in a differentiable search process. This enables end-to-end training of the model from human demonstrations. (b) During the planning phase, the estimator generates node-wise relaxation costs, and a discrete graph search process selects regions to relax, inducing an admissible space to facilitate path planning.
    } 
    \label{fig:architecture}
  \vspace{-1.5em}    
\end{figure*}

\section{Method}

We introduce our interleaving planning and execution with regional constraint relaxation in semi-static environments.

\subsection{Problem formulation}\label{ssec_formulation}
Consider a navigation space $\mathcal{R}$ partitioned into regions with no, soft, and hard constraints: $\mathcal{R}=\mathcal{R}_{\text{free}} \cup \mathcal{R}_{\text{soft}}  \cup \mathcal{R}_{\text{hard}}$. We aim to find a safe and efficient route $\mathcal{X}=\{ \vx(s) | s\in [0,1]\}$, where $\vx(s)$ is a configuration at the phase $s$, that traverses the least-critical (i.e., relaxed) region $\mathcal{R}_{\text{rlx}}\subseteq \mathcal{R}_{\text{soft}}$, yielding the admissible set $\mathcal{R}_{\text{safe}}= \mathcal{R}_{\text{free}} \cup \mathcal{R}_{\text{rlx}} $. We formulate the route search and relaxation problem as 
\begin{align}
&\argmin_{\mathcal{X}, \mathcal{R}_{\text{rlx}}\subseteq \mathcal{R}_{\text{soft}}} && \int_{0}^1 \| \dot{ \vx }(s) \| ds + \Psi_{\text{cost}}(\mathcal{R}_{\text{rlx}}) \\
&\text{such~that} && \vx(s)\in \mathcal{R}_{\text{free}} \cup \mathcal{R}_{\text{rlx}} \;\; \forall s \in[0,1], \nonumber \\
& && \vx(0)=\text{start}, \vx(1)=\text{goal}, \nonumber
\label{eq_problem}
\end{align}
where $\Psi_{\text{cost}}$ assigns a non-negative relaxation penalty cost ($\in\mathbb{R}^+$) to traversing the relaxed region $\mathcal{R}_{\text{rlx}}$. This problem is generally NP-hard due to the non-additive coupling between the continuous length and the discrete choice of relaxations.  

To approximate the problem, we reformulate Eq.~(\ref{eq_problem}) leveraging a graph-based region abstraction. The graph $G=(\mathcal{V}, \mathcal{E})$ consists of region nodes $\mathcal{V}=\{\vr_i\}_{i=1}^N$ and region-adjacency edges $\mathcal{E}=\{(\vr_i, \vr_j) | i\neq j,  \mathcal{H}(\vr_i, \vr_j ) \geq \tau  \}$ where each node $\vr_i \subset \mathcal{R}_{\text{free}}\cup\mathcal{R}_{\text{soft}}$ denotes a spatial segment with centroid $c(\vr_i)\in\mathbb{R}^2$, $\mathcal{H}$ measures boundary affinity, and $\tau$ is the affinity threshold for local motion. We then project the continuous path $\mathcal{X}$ onto $G$ as a sequence of region memberships:
\begin{align}
\mathcal{X}_G=[\vr_{\mathcal{M}(1)}, ..., \vr_{\mathcal{M}(|\mathcal{X}|)}],\; \mathcal{X}[t]\in \vr_{\mathcal{M}(t)},
\end{align}
where $t\in [1, |\mathcal{X}|]$ and $\mathcal{M}:\{1, ..., |\mathcal{X}|\}\rightarrow  \{1, ..., N\}$, which maps each configuration $\mathcal{X}[t]$ to its containing region node. Equivalently, the projected path connects centroids $\{c(\vr_{\mathcal{M}(t)})\}_{t=1}^{|\mathcal{X}|}$ on the graph map. We then represent the graph-based optimization problem as
\begin{align}
\label{eq_new_problem}
&\argmin_{\mathcal{X}_G} && \sum_{t=2}^{|\mathcal{X}_G|} d(\vr_{\mathcal{M}(t-1)}, \vr_{\mathcal{M}(t)}) \nonumber \\
& && \;\;\; +\sum_{t=1}^{|\mathcal{X}_G|}\mathds{1} \left[ \vr_{\mathcal{M}(t)}\in\mathcal{R}_{\text{rlx}}\right] \Psi_{G}(\vr_{\mathcal{M}(t))})  \\
&\text{such~that} && \vr_{\mathcal{M}(t)}\in \mathcal{R}_{\text{free}} \cup \mathcal{R}_{\text{rlx}} \;\; \forall t \in[1,|\mathcal{X}_G|], \nonumber \\
& && (\vr_{\mathcal{M}(t-1)}, \vr_{\mathcal{M}(t)})\in\mathcal{E} \;\; \forall t \in[1, |\mathcal{X}_G|], \nonumber \\
& && \vr_{\mathcal{M}(1)}=\vr_\text{start}, \vr_{\mathcal{M}(|\mathcal{X}_G|)}=\vr_\text{goal}, \nonumber \\
& && \mathcal{R}_{\text{rlx}} = \mathcal{R}_{\text{soft}} \cap \mathcal{X}_G, \nonumber  
\end{align}
where $d$ is a geodesic distance measure between the center of two nodes and $\Psi_{G}$ is a relaxation cost function on the graph $G$. For notational convenience, we denote the objective function in Eq.~(\ref{eq_new_problem}) by $\mathcal{L}_{\text{search}}$. %

Fig.~\ref{fig:architecture} shows the overall architecture of our method. It optimizes navigation by constructing a superpixel-based graph $G$, training $\Psi_{G}$, and applying a differentiable A* algorithm (see Sec.~\ref{ssec_relaxation}). The process consists of 1) superpixel graph construction (Sec.~\ref{ssec_superpixel}), 2) graph-based path planning with constraint relaxation (Sec.~\ref{ssec_relaxation}), and 3) reactive planning that interleaves path search and constraint relaxation (Sec.~\ref{sec_reactive}). 

\subsection{Superpixel Graph Generation with Semantic Labels}\label{ssec_superpixel}
We construct a superpixel graph map $G=(\mathcal{V}, \mathcal{E})$ for human-like relaxation region selection and path planning, where $\mathcal{V}$ and $\mathcal{E}$ denote superpixel-based node regions and region-adjacency edges, respectively. Superpixels partition an image-based map into over-segmented regions that do not cross object boundaries~\cite{stutz2018superpixels}. By constraining pixels within the same region, superpixels enable homogeneous semantic-labels, improving prediction consistency and sharpening relaxation-region boundaries. Further, their compact graph representation offers controllable granularity with near-uniform regions that support stable planning \cite{chen2025constraint}.

To construct $G$, we partition the traversable space $\mathcal{R}_{\text{free}}\cup \mathcal{R}_{\text{soft}}$ into $N$ node regions $\mathcal{V}=\{\vr_i\}_{i=1}^N$ using SLIC~\cite{achanta2010slic}\textemdash a K-Means based pixel clustering algorithm. Note that all locations within each node region $\vr_i$ share the same semantic label.
Each node $\vr_i\in\mathcal{V}$ contains a pair of vectors $[\vf_{\text{GR}}, \vf_{\text{SR}}]_{\vr_i}$:
\begin{itemize}[leftmargin=*]
\item $\vf_{\text{GR}}$ contains geometric information, the center coordinate $c(\vr_i)\in\mathbb{R}^2$ for distance computation in planning.
\item $\vf_{\text{SR}}$ contains semantic information, including the semantic label, a start-region indicator, and a goal-region indicator of the current node $\vr_i$:
\begin{align}
\vf_{\text{SR}}=\left[ \Phi_{\text{label}}(\vr_i), \mathds{1}[\vx_{\text{start}}\in\vr_i], \mathds{1}[\vx_{\text{goal}}\in\vr_i] \right],
\end{align}
where $\Phi_{\text{label}}: \mathcal{V} \rightarrow \mathbb{B}^{M}$ is a semantic classifier returning a one-hot vector of dimension $M$. 
\end{itemize}
Each edge $e_{ij}\in\mathcal{E}$ contains a distance measure between two geometrically adjacent regions; $e_{ij}=d(\vr_i, \vr_j)$.

After the initial construction of $G$, we incrementally update the graph by re-segmenting each region with new observations and re-constructing edges. This update mechanism maintains the label consistency and compactness of the graph allowing interleaving planning and execution.

\subsection{Graph-based Path-Planning and Relaxation}\label{ssec_relaxation}
We present a GNN-based regional constraint-relaxation method that couples a neural relaxation-cost estimator with a differentiable A* planner to generate preferable paths. To imitate human-like relaxed path generation, we learn the relaxation penalty $\Psi_{G}$ in Eq.~(\ref{eq_new_problem}) from human navigation demonstrations collected in over-constrained environments so that the planner emulates human-like relaxation behavior. Below, we describe the demonstration data and the two learning stages in detail.

\noindent\textbf{1) Expert demonstrations}: We collect a $N_D$ number of expert demonstrations $\mathcal{D}=\{^\mathcal{D}\mathcal{X}^{(i)} \}_{i=1}^{N_D}$, where each sequence $^\mathcal{D}\mathcal{X}^{(i)}=[\vx(0), ..., \vx(1) ]$ is a sequence of 2-D coordinates. Each path illustrates how a human expert simultaneously traces a continuous path and, when necessary, passes through one or more regions labeled with soft constraints $\mathcal{R}_{\text{soft}}$. We then project the demonstrations $\mathcal{D}$ onto the graph map $G$ by converting each path $^\mathcal{D}\mathcal{X}^{(i)}$ into $^\mathcal{D}\mathcal{X}^{(i)}_G=[\vr_{\mathcal{M}(0)}, ..., \vr_{\mathcal{M}(|^\mathcal{D}\mathcal{X}^{(i)}|)}]$ where $^\mathcal{D}\mathcal{X}^{(i)}[t]\in\vr_{\mathcal{M}(t)}$. The data collection detail is available in Sec.~\ref{sec_setup}.

\noindent\textbf{2) Relaxation cost estimation}: We model a neural relaxation-cost estimator $\Psi_{G}: \mathcal{V} \rightarrow \mathbb{R}^+$ to find a path balancing the path length and relaxation penalty via a differentiable graph-based A* planner. The estimator is a GNN model using a superpixel graph $G_s=(\mathcal{V}_s, \mathcal{E})$, which is the priorly constructed $G$ with a part of node features, $\vf_{\text{SR}}$. As shown in Fig.~\ref{fig:architecture}, the model consists of a stack of GPS layers \cite{rampavsek2022recipe}, where each layer is a hybrid layer of a message-passing neural network (MPNN) and a global attention network that captures long-range dependency in a graph. Following GraphGPS \cite{rampavsek2022recipe}, we use GatedGCN \cite{bresson2017residual} as the MPNN layer and Transformer \cite{vaswani2017attention} as the global attention layer. 

The GPS layers update encoded node and edge features for the node-wise cost estimation. Let $\vV^k \in \mathbb{R}^{N\times (M+2)}$ and $\vE^k \in \mathbb{R}^{|\mathcal{E}|}$ be the encoded node features and edge features, respectively. The updated process is
\begin{align}
\vV^{k+1}, \vE^{k+1} = \mathrm{GPS}^{k} (\vV^{k}, \vE^{k}, \vA),
\end{align}
where $\vA\in\mathbb{R}^{N\times N}$ is the adjacency matrix of the graph $G_s$ and $k\in\{1, ... , K\}$ is the number of GPS layers. Finally, applying a node-wise shared multi-layer perceptrons (MLPs) with a softplus activation, our cost estimator outputs node-wise scores $\{\Psi_{G}(\vr_i)\}_{i=1}^{N}$. Note that we zero-mask the scores for nodes in free space $\vr_i\in\mathcal{R}_{\text{free}}$ not to assign relaxation cost for non-constrained regions.

\noindent\textbf{3) GNN-based Planning with Relaxation Cost}: 
Our method searches for an optimal path that may traverse constrained regions while explicitly accounting for relaxation costs via differentiable graph-basd A* planner. We incorporate these costs by augmenting the accumulated path costs to 
\begin{align}
\mathbf{g}^{\text{ext}}=\mathbf{g}+[\Psi_{G}(\vr_i)]_{i=1}^N.
\end{align}
The extended costs modify the planner's selection rule, replacing Eq.~(\ref{eq_v_sel}) as
\begin{align}
\mathbf{v}_{\text{sel}} = I_{\text{max}}\left(\frac{\exp(-(\mathbf{g}^{\text{ext}}+\mathbf{h})/\lambda)\odot\mathbf{o}}{\exp(-(\mathbf{g}^{\text{ext}}+\mathbf{h})/\lambda)\mathbf{o}}\right).
\end{align}
We then add the region entry penalty of the selected node parent by redefining Eq.~(\ref{eq_g_prime}) as
\begin{align}
\mathbf{g}'&=\mathbf{g}^{\text{ext}}\odot\mathbf{v}_\text{sel}+\mathbf{W}\mathbf{v}_\text{sel}.
\end{align}
As described in Sec.~\ref{sec_preliminary}, these modifications allow differentiable graph A* to find the minimum-cost path $\mathcal{X}^*$ under relaxation penalties across $\mathcal{R}_{\text{free}}\cup \mathcal{R}_{\text{soft}}$.

\subsection{End-to-End Training}\label{ssec_training}
We learn the relaxation cost estimator $\Psi_{G}$ from human demonstration $\mathcal{D}$ by leveraging the differential A* planner. As shown in Fig.~\ref{fig:architecture}, the differentiable search module propagates the relaxation error that we will describe in Eq.~(\ref{relax_loss})
backward enabling end-to-end training of the cost estimator $\Psi_{G}$. We further update $\Psi_{G}$ introducing a mini-batch loss $\mathcal{L}_{\text{batch}}$ to minimize false positive and negative relaxations. 

Consider a batch $\mathcal{B}\subset\mathcal{D}$. The false region relaxation loss is a measure of how much the A* planner traverses the regions, with soft constraints, that were not explored in each demonstration ${^\mathcal{B}\mathcal{X}^{(i)}}$. The A* planner yields an explored constraint region mask $\mathbf{v}_{\text{rlx}}^{(i)}\in\mathbb{B}^N$ that is the intersection of the closed set $\vc$ and $\mathcal{R}_{\text{soft}}$. From each demonstration, we also extract a mask $^\mathcal{B}\mathbf{v}_{\text{rlx}}^{(i)}\in\mathbb{B}^N$ for relaxed regions with constraints in $\mathcal{R}_{\text{soft}}$. Then, we measure the false relaxation as the weighted sum of false positives and negatives,
\begin{align}
&\mathcal{L}_{\text{rlx}}(\mathbf{v}_\text{rlx}, {\mathbf{v}_\text{rlx}^{*}})\! = \!\frac{w_\text{fp} \vv_\text{rlx}^\top(\mathbf{1}\!\!-{\vv_{\text{rlx}}^{*}})+w_\text{fn} (\mathbf{1}\!\!-\vv_\text{rlx})^\top {\vv_{\text{rlx}}^{*}} }{1+\|{\mathbf{v}_{\text{rlx}}^{*}} \|_1} 
\label{relax_loss}
\end{align}
where $w_{\text{fp}}$ and $w_{\text{fn}}$ are hyper parameters ($\in \mathbb{R}$), and the denominator is for normalization. Note that the A* search tends to increase false positives so that we set $w_\text{fn}>w_\text{fp}$ to reduce sample bias.

We finally formulate a mini-batch loss $\mathcal{L}_{\text{batch}}$ with per-sample weight $w^{(i)}$ as
\begin{align}
\mathcal{L}_{\text{batch}} &= 
\frac{1}{|\mathcal{B}|}\sum_{i=1}^{|\mathcal{B}|} w^{(i)} \mathcal{L}_{\text{rlx}}(\mathbf{v}_{\text{rlx}}^{(i)},{^{\mathcal{B}}\mathbf{v}_{\text{rlx}}^{(i)}}), \\
w^{(i)} &= \min\left(1, \mathcal{L}_{\text{rlx}}(\mathbf{v}_\text{rlx}^{(i)}, {^{\mathcal{B}}\mathbf{v}_\text{rlx}^{(i)}})^\gamma \right) 
\end{align}
where $\gamma > 0$. In this work, we use another set of hyper parameters, $w_{\text{fp}}=0.5$ and $w_{\text{fn}}=0.5$, for the per-sample weight $w^{(i)}$.

\section{Interleaving Planning and Execution}\label{sec_reactive}
\begin{algorithm}[t]
\small
\caption{Interleaving planning and execution with constrained region relaxation. $\mathrm{O}_t$ is the sensor observation at time step $t$.}
\label{alg:online-relax-ema-min}
\raggedright
\textbf{Inputs} $\vx_{\text{start}}$, $\vx_{\text{goal}}$, $G_0$ \\

\begin{tabbing}
\hspace*{1.1em}\=\hspace*{1.0em}\=\kill
$t\!\gets\!1, \mathcal{R}_{\text{free}, 0}\!\gets\! \emptyset, \mathcal{R}_{\text{soft}, 0}\!\gets\! \emptyset, \mathcal{R}_{\text{hard}, 0} \!\gets\! \emptyset$ \\
\textbf{while} true \textbf{do} \\
\> $\mathcal{R}_{\text{free}, t}, \mathcal{R}_{\text{soft}, t}, \mathcal{R}_{\text{hard}, t}  \!\xleftarrow[]{\text{update}}\! (\mathcal{R}_{\text{free}, t-1}, \mathcal{R}_{\text{soft}, t-1}, \mathcal{R}_{\text{hard}, t-1}, \mathrm{O}_t)$ \\
\> $G_t \!\xleftarrow[]{\text{update}}\! (G_{t-1}, \mathcal{R}_{\text{free}, t}, \mathcal{R}_{\text{soft}, t})$ \quad\quad \% $G_t=(\mathcal{V}_t, \mathcal{E}_t)$\\
\> $\mathbf{g}^{\text{ext}} \!\gets\! \mathbf{g}+\Psi_{G}(\mathcal{V}_t)$ \\
\> $\mathcal{X}_G \!\gets\! \textsc{A*}(G_t,\vx_\text{start},\vx_\text{goal}, \mathbf{g}^{\text{ext}})$ \quad\quad\quad \% \text{Graph-based search}\\
\> $\mathcal{R}_{\mathrm{rlx}} \!\gets\! \mathcal{R}_{\text{soft}, t} \cap \mathcal{X}_G $\\
\> $\mathcal{X} \!\gets\! \textsc{A*}(\mathcal{R}_\text{free}\cup\mathcal{R}_\text{rlx}, \mathbf{x}_\text{start}, \mathbf{x}_\text{goal})$ \quad\quad \% \text{Fine grid-based search} \\
\> $\mathrm{O}_{t+1} \!\gets\! \textsc{Execute}(\mathcal{X})$ \\
\> \textbf{if} \textsc{Reached}$(\mathbf{x}_\text{start},\mathbf{x}_\text{goal})$ \textbf{or} \textsc{Terminated}() \textbf{then break} \\
\> $t \!\gets\! t{+}1$
\end{tabbing}
\vspace{-0.5em}
\end{algorithm}

We introduce an interleaving planning and relaxation framework to achieve asymptotic completeness in semi-static environments. Algorithm~\ref{alg:online-relax-ema-min} shows the pseudo code of the interleaving planning and execution framework while relaxing constrained region according to the relaxation cost-based graph-path search until it reaches a goal. Our algorithm updates the navigation space $\mathcal{R}_t$ and corresponding graph $G_t$ upon new observation at each time step $t$. Leveraging an A* planner with the proposed relaxation cost estimator $\Psi_{G}$, our method finds the graph-based safe and efficient path from the start $\vx_\text{s}$ to the goal $\vx_\text{g}$. Note that we assume the continuous space start and goal to be projected to the region nodes on the graph internally. 

In low-level planning, our framework provides multiple discretizations and coverages of path plans in the continuous navigation space $\mathcal{R}$. 
\begin{itemize}[leftmargin=*]
\item \textbf{Graph plan}: By default, we produce a best-effort graph plan $\mathcal{X}_G$ which is a sequence of regions. The robot tracks the centroid waypoints $\{c(\mathcal{X}_G[t])\}_{t=1}^{|\mathcal{X}_G|}$. In practice, the centroid-to-centroid motion between adjacent regions may not be locally feasible, we optionally extract a portal-sequence with shared boundary segments and plan local motions between consecutive portals. 
\item \textbf{Continuous plan within $\mathcal{X}_G$}: We refine the route by computing a continuous path $\mathcal{X}$ restricted to $\mathcal{X}_G$, using a finer-grid search (e.g., standard A*) or a local continuous planner. This enforces local feasibility and improves smoothness while respecting the region selection.
\item \textbf{Continuous plan in $\mathcal{X}_G\cup \mathcal{R}_{\text{free}}$}: We further allow $\mathcal{X}$ to traverse both the planned regions and nearby free space, seeding the search with $\mathcal{X}_G \cup \mathcal{R}_{\text{free}}$. A finer-grid planner in this unified space yields shorter and safer trajectories while preserving the high-level guidance of the graph plan.
\end{itemize}
In this work, we use the continuous plan in $\mathcal{X}_G\cup \mathcal{R}_{\text{free}}$.

\section{Experimental Setup}\label{sec_setup}

\subsection{Quantitative Evaluation}
We create a semi-static navigation benchmark to evaluate the human-likeness and the safety\textendash efficiency trade-off. The benchmark consists of $34$\ $\SI{1}{\kilo\meter}\times\SI{1}{\kilo\meter}$ OpenStreetMap~\cite{OpenStreetMap} tiles sampled from global cities and aligned with satellite imagery.
We split them into $26$ for training, $5$ for validation, and $3$ for testing. We convert each tile into a 2D semantic bird-eye-view map using MIA~\cite{ho2024map}, with up to $10$ semantic labels: 1) \textit{sidewalk} for $\mathcal{R}_{\text{free}}$, 2) \textit{crosswalk}, \textit{road}, \textit{living street}, \textit{parking lot}, \textit{grass}, and \textit{rough terrain} for $\mathcal{R}_{\text{soft}}$, and 3) \textit{building} and \textit{water} for $\mathcal{R}_{\text{hard}}$. 

For training and validation, we construct $1,240$ navigation environments by randomly sampling $20$ start-goal pairs per map and creating two variants for each: one with static conditions and one with semi-static conditions. We constrain start-goal pairs to \textit{sidewalks} and space them \SI{30}{\meter}-\SI{200}{\meter} apart. We collect one demonstration per environment, yielding $1,240$ demonstrations. For each static environment, a human expert draws a trajectory using a mouse. To synthesize the corresponding semi-static variant, we randomly sample a superpixel seed along the trajectory, alter neighboring superpixels to obstruct the pathway, and ask the expert to redraw the trajectory from that point, reusing the prefix up to the perturbed superpixel. This results in $620$ static and $620$ semi-static demonstrations. 

To evaluate the safety\textendash efficiency trade-off, we construct $300$ environments from three diverse test maps\textemdash \textrm{Milan}, \textrm{Baltimore}, and \textrm{Capitol Hill}\textemdash by randomly sampling $100$ start-goal pairs per map. We perturb them sampling up to three superpixel seeds along the Open-Source Routing Machine (OSRM)~\cite{luxen-vetter-2011}~-generated subgoal paths, with all changes aggregated per map to increase disruption likelihood. Each method plans on the original map but navigates in the modified one. For human-likeness evaluation, we generate $100$ semi-static environments with demonstrations from five test maps, including two addional maps, sampling $20$ start-goal pairs per map.

We evaluate SuReNav against all baselines on 2D maps, except ViPlanner~\cite{roth2024viplanner}, which requires visual input. 
For ViPlanner, we reconstruct the 2D maps in the Isaac Lab 3D simulator~\cite{mittal2023orbit}, manually refining OpenStreetMap nodes to preserve pedestrian topology and routing consistency. Below, we detail the baselines:

\begin{figure*}[t]
    \centering
    \includegraphics[width=\linewidth]{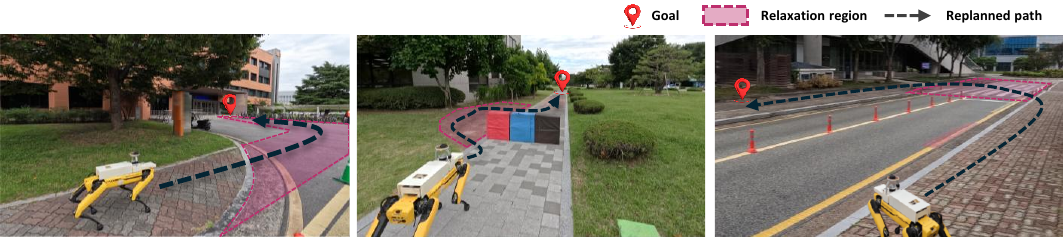}
    \caption{ Real-world evaluation in semi-static settings. The quadruped robot begins by following a path planned using an outdated prior map. Upon encountering newly constrained regions, SuReNav dynamically relaxes a red-colored soft-constraint region and adapts its behavior to perform human-like navigation that balances safety and efficiency. Video demonstrations are available on the project website.}
    \label{fig:real_scenario}
    \vspace{-1em}
\end{figure*}

\begin{itemize}[leftmargin=*, noitemsep,topsep=0pt]
\item \textbf{ViPlanner\textsuperscript{OSRM}}~\cite{roth2024viplanner,luxen-vetter-2011}: A vision-based local navigation approach extended from ViPlanner~\cite{roth2024viplanner}. Due to the local nature, ViPlanner requires subgoals to reach distant targets. To address this, we integrate ViPlanner with OSRM to provide global pedestrian routes as subgoal sequences. If consecutive subgoals exceed \SI{15}{\meter}, we insert intermediate subgoals to satisfy ViPlanner’s goal specification. We also align ViPlanner’s semantic label table with the region priorities and costs used in other baselines.
\item \textbf{D* Lite}~\cite{koenig2002d}: D* Lite is an incremental heuristic search algorithm that replans by repairing the shortest path as edge costs change. We derive traversal costs from the reciprocal proportion of demonstration-traversed superpixels with $0$-$1$ scaling, enabling pixel-level trajectory optimization under the evolving cost field.
\item \textbf{COA*}~\cite{lim2021generalized}: Class-Ordered A* (COA*) is a generalized extension of A* that incorporates hierarchical class ordering into its search process. We determine the class priority using the traversal frequency observed in the expert demonstration data, allowing COA* to relax constraints that respect a fixed, data-driven priority order.
\item \textbf{Rule-based Constraint Relaxation (RCR)}~\cite{kim2024reactive}: We denote this unnamed method as RCR, a reactive framework that relaxes constraints by manually scoring regions based on predefined label-to-risk mappings and distances to the goal.
We also replace our learned cost estimator with RCR's scoring function to compare our learning-based approach and predefined heuristic-based strategy.
\end{itemize}
\smallskip

We assess human-likeness by comparing planned trajectories with expert demonstrations focusing on geometric similarity and constraint violations using the following metrics:
\begin{itemize}[leftmargin=*, noitemsep,topsep=0pt]
    \item \textbf{Fréchet distance}: Distance between two continuous curves~\cite{eiter1994computing}, with each sample's Fréchet distance normalized by the Euclidean distance between start and goal.
    \item \textbf{Relaxation IoU}: Intersection-over-Union (IoU) between predicted and ground-truth relaxed region sets.
\end{itemize}

\smallskip
We then benchmark using the following metrics:
\begin{itemize}[leftmargin=*, noitemsep,topsep=0pt]
    \item \textbf{Success Rate (SR)}: Proportion of episodes in which the robot reaches its goal without entering any hard-constrained regions, $\mathcal{R}_{\text{hard}}$.
    \item \textbf{Success weighted by path length (SPL)}: Efficiency metric comparing the executed path length to the shortest feasible path length, weighting zero to failed episodes. When robot reaches its goal without entering any hard-constrained regions $\mathcal{R}_{\text{hard}}$, the episode is noted as success.
    \item \textbf{Total Risk}: Cumulative per pixel risk along a trajectory, where each label's unit risk is the mean $0$-$1$ scaled score from three commercial large language models\textemdash ChatGPT, Gemini, and Claude. This metric captures both risk magnitude and exposure length for assessing safety.
\end{itemize}

\begin{table}[t]
  \centering
  \caption{Comparison of human-likeness to expert demonstrations. We compute Fréchet distance and relaxation IoU metrics against $100$ expert demonstration trajectories from five unseen global city maps.}
  \label{tab:humanlikeness}
  \begin{tabular}{l
                  c 
                  c}
    \toprule
    \multirow{1}{*}{Method}
      & {Fréchet distance $(\downarrow)$}
      & {Relaxation IoU $(\uparrow)$} \\
    \midrule
    D* Lite   & 0.429  & 0.223 \\
    COA*     & 0.701 & 0.271 \\
    RCR        & 0.626 & 0.259 \\
    SuReNav (Ours)        &\textbf{0.334} &\textbf{0.416} \\
    \bottomrule
  \end{tabular}
  \vspace{-1em}
\end{table}
\begin{figure}[t]
    \centering
     \includegraphics[width=\linewidth]{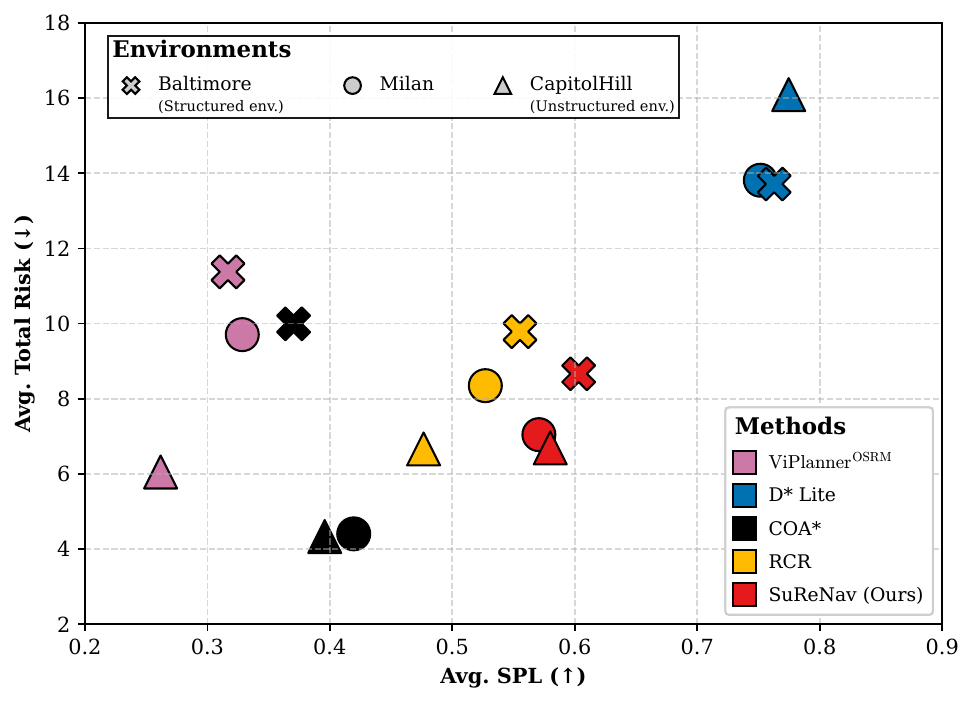}
    \caption{Comparison of the proposed SuReNav and four baseline method in $300$ simulated navigation environments from three urban areas\textemdash \textrm{Milan}, \textrm{Baltimore}, and \textrm{Capitol Hill}. We evaluate average SPL and Total Risk metrics where performance improves toward the lower-right corner that is higher SPL and lower Total Risk.}
    
    \label{fig:SPL_RISK_result}
        \vspace{-1em}
\end{figure}
\begin{figure*}[t]
 \centering
  \includegraphics[width=\textwidth]{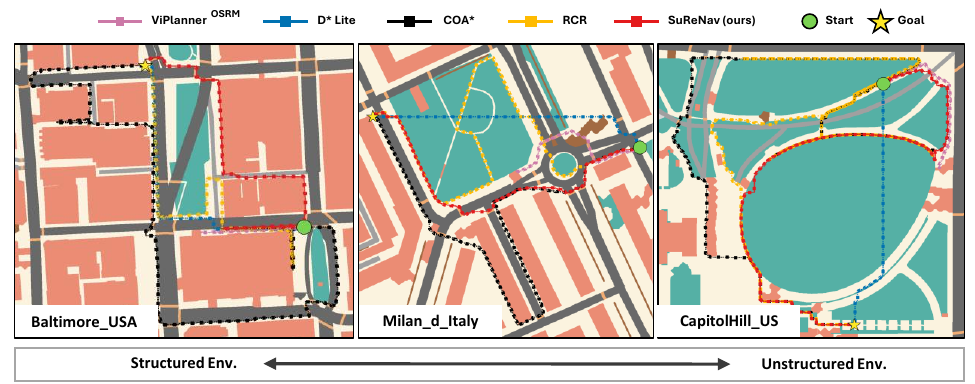}
  \caption{Examples of navigation behaviors across three urban scenarios. Green circles and yellow stars denote start and goal locations, respectively. The ivory, green, light orange, gray, orange, and blue regions correspond to \textit{sidewalks}, \textit{lawns}, \textit{crosswalks}, \textit{roads}, \textit{buildings}, and \textit{water} areas, respectively. Colored line segments indicate the trajectories of individual navigation examples.}
  \label{fig:simulation-examples}
\end{figure*}

\begin{figure*}[t]
    \centering
    \includegraphics[width=\linewidth]{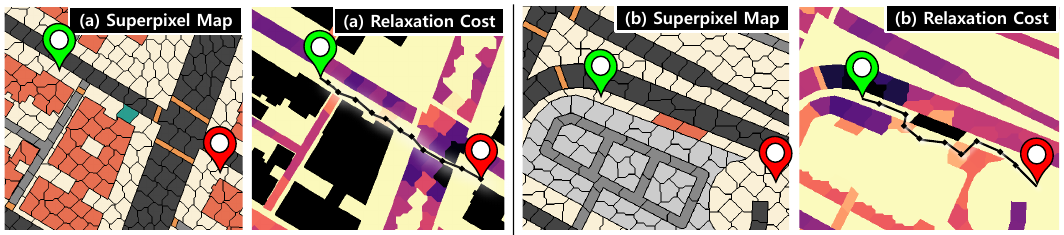}
    \caption{Visualization of learned relaxation cost distributions in two distinct navigation scenarios. The ivory, green, light orange, light gray, gray, dark gray, and orange regions in the superpixel maps correspond to \textit{sidewalks}, \textit{lawns}, \textit{crosswalks}, \textit{parking lots}, \textit{living streets}, \textit{roads} and \textit{buildings}, respectively. Brighter colors indicate lower relaxation costs, marking regions as suitable for traversal during planning. In (a), SuReNav relaxes a grass region to create an efficient shortcut, while in (b), it navigates around a newly introduced obstacle by identifying the adjacent parking lot as the optimal path for relaxation.}
    \label{fig:cost_distribution}
    \vspace{-1em}
\end{figure*}

\subsection{Qualitative Evaluation}
We demonstrate our method in campus environments on a quadruped robot, Boston Dynamics Spot, equipped with an Intel Realsense D435i camera and an $128$-channel Ouster LiDAR. The robot constructs a superpixel region-adjacency graph by performing PIDNet-based semantic segmentation~\cite{xu2023pidnet} and aggregating pixel labels into regions. To emulate semi-static conditions, we inject regional constraints such as adding obstacles during navigation or obfuscating portions of the prior map before execution as shown in Fig.~\ref{fig:real_scenario}.

\section{Evaluation}
We evaluate the human-likeness of relaxation-based navigation against $100$ expert demonstrations in semi-static environments. As shown in Table~\ref{tab:humanlikeness}, SuReNav achieves the smallest Fréchet distance to the demonstrations while relaxing a set of regions that closely matches human selections. Notably, its relaxation IoU is $44\%$ higher than the next-best approach COA* since all baselines rely on pre-defined, fixed regional costs. In contrast, SuReNav predicts context-conditioned regional costs from demonstrations, enabled by the graph-based relaxation cost estimator that jointly accounts for nearby risky regions and navigation context. This enables SuReNav to effectively learn human navigation behaviors in semi-static environments. Note that we exclude ViPlanner\textsuperscript{OSRM}, as it fails to generate complete subgoals connecting the start or goal of human demonstrations due to the limited distribution of OSRM nodes.

Fig.~\ref{fig:SPL_RISK_result} compares SuReNav with four baselines in simulated semi-static environments; higher SPL and lower RPM indicate better performance. SuReNav achieves comparable success and efficiency while producing lower-risk trajectories across all structure levels of environments. COA* occasionally yields lower risk in unstructured cities\textemdash Milan and Capitol Hill\textemdash by minimizing the number of constraint violations, but its trajectories are about $50\%$ longer than SuReNav's. D* Lite attains the highest SPLs, yet its reliance on fixed regional costs leads to roughly twice the Total Risk in unseen settings. In contrast, SuReNav does not require manual cost specification and generalizes through learning from demonstrations. Note that all methods achieve $100\%$ SR except ViPlanner, which records $51\%$.

Fig.~\ref{fig:simulation-examples} illustrates exemplar navigation behaviors in three urban environments. In the Baltimore and Milan maps, SuReNav efficiently reaches the goals using only \textit{sidewalks} and \textit{crosswalks}, avoiding jaywalking or traversing \textit{lawns}. In detail, as shown in Fig.~\ref{fig:cost_distribution}, SuReNav creates safe and efficient paths by assigning low costs only to specific superpixels on $\mathcal{R}_{\text{soft}}$, enabling the formation of complete paths, while assigning high costs to the surrounding areas to guide against unexpected violations. In contrast, D* Lite often traverses \textit{roads} or \textit{lawns} to produce shorter paths in unstructured city setups, as its incremental local search fails to account for accumulated risks along the route. On the other hand, COA* exhibits the longest trajectories in both environments since COA* prioritizes minimizing the number of constraint relaxations, thus producing the most conservative routes.

We finally demonstrated SuReNav in real-world campus navigation scenarios using the quadruped robot Spot. Fig.~\ref{fig:real_scenario} presents three examples of human-like constraint-relaxation behaviors. In one case, the robot crosses a \textit{road} at the end of a \textit{sidewalk} by relaxing high-risk \textit{road} regions (red), mirroring human navigation strategies. In another example (middle), SuReNav relaxes only a small portion of a constrained region instead of traversing a large area, demonstrating precise and selective relaxation. Finally, as shown on the right, SuReNav directs Spot to take a distant crosswalk, sacrificing efficiency to ensure safety. This emergent behavior, learned from demonstrations rather than manual programming or cost defining, highlights the framework's ability to balance efficiency and adherence to traffic rules in novel environments. Experiment videos are available on the project website.

\section{Conclusion}
This paper introduced SuReNav, a constraint-aware framework that interleaves relaxation, planning, and execution to balance safety and efficiency in semi-static environments. SuReNav learns human-like relaxation and planning behaviors from expert demonstrations, formalizing the problem with superpixel-graph map construction, regional relaxation-cost estimation, and relaxed path search by jointly training a relaxation cost estimator and a differentiable graph-based A* planner. The estimator treats trivially safe areas (e.g., \textit{sidewalks}) as unconstrained, while handling other traversable regions as soft constraints\textemdash initially forbidden but admissible upon explicit relaxation. Through extensive evaluations in simulated environments, we show that SuReNav produces human-like, safe, and efficient navigation in novel settings. Finally, its real-world deployment on a quadruped robot validates its generalization beyond simulation.

\addtolength{\textheight}{-12cm}   %

\bibliographystyle{ieeetr}
\bibliography{references}

\end{document}